\documentclass{bmvc2k}

\title{$\boldsymbol M^2KD$: Incremental Learning via Multi-model and Multi-level Knowledge Distillation}

\addauthor{Peng Zhou}{pengzhou@umd.edu}{1}
\addauthor{Long Mai}{malong@adobe.com}{2}
\addauthor{Jianming Zhang}{jianmzha@adobe.com}{2}
\addauthor{Ning Xu}{nxu@adobe.com}{2}
\addauthor{Zuxuan Wu}{zxwu@cs.umd.edu}{1}
\addauthor{Larry S. Davis}{lsd@umiacs.umd.edu}{1}
\addinstitution{
University of Maryland\\
College Park, MD. USA.\\
}
\addinstitution{
Adobe Research\\
345 Park Avenue\\
San Jose, CA. USA.\\
}

\runninghead{Zhou \lowercase{et al.}}{Incremental Learning via $\boldsymbol M^2KD$}

\def\eg{\emph{e.g}\bmvaOneDot}

\def\etal{\emph{et al}\bmvaOneDot}
\usepackage{algpseudocode,algorithm,algorithmicx}

\begin{document}

\maketitle

\begin{abstract}
Incremental learning targets at achieving good performance on new categories without forgetting old ones. Knowledge distillation has been shown critical in preserving the performance on old classes. Conventional methods, however, sequentially distill knowledge only from the penultimate model, leading to performance degradation on the old classes in later incremental learning steps. In this paper, we propose a multi-model and multi-level knowledge distillation strategy. Instead of sequentially distilling knowledge only from the penultimate model, we directly leverage all previous model snapshots. In addition, we incorporate an auxiliary distillation to further preserve knowledge encoded at the intermediate feature levels. To make the model more memory efficient, we adapt mask based pruning to reconstruct all previous models with a small memory footprint. Experiments on standard incremental learning benchmarks show that our method improves the overall performance over standard distillation techniques.
\end{abstract}

\section{Introduction}

Deep neural networks perform well on many visual recognition tasks~\cite{girshick2015fast,long2015fully,krizhevsky2012imagenet} given specific training data. However, problem arises when adapting networks to unseen categories while remembering seen ones, which is known as catastrophic forgetting~\cite{mccloskey1989catastrophic,goodfellow2013empirical,kirkpatrick2017ewc}. To tackle this issue, there is a growing research attention on incremental learning where the new training data is not provided upfront but added incrementally. The target of incremental learning is to achieve good performance on new data without sacrificing the performance on old and it has been widely explored across different tasks such as classification~\cite{li2018lwf,rebuffi2017icarl} and detection~\cite{shmelkov2017object}.

To alleviate catastrophic forgetting in incremental learning, 
one possibility is to maintain a subset of old data to avoid over fitting on new data~\cite{rebuffi2017icarl,Castro2018end2end,li2018supportnet,wu2019BiC}. However, an issue in practice is that when models embedded in a product are delivered to customers, they no longer have access to trained data for privacy purposes. To tackle the situation, a stricter exemplar-free setting was introduced in~\cite{li2018lwf}, which requires no exemplar set for previous categories and only distills previous knowledge from the current categories.

\begin{figure}[t]
\begin{center}
   \includegraphics[width=0.7\linewidth]{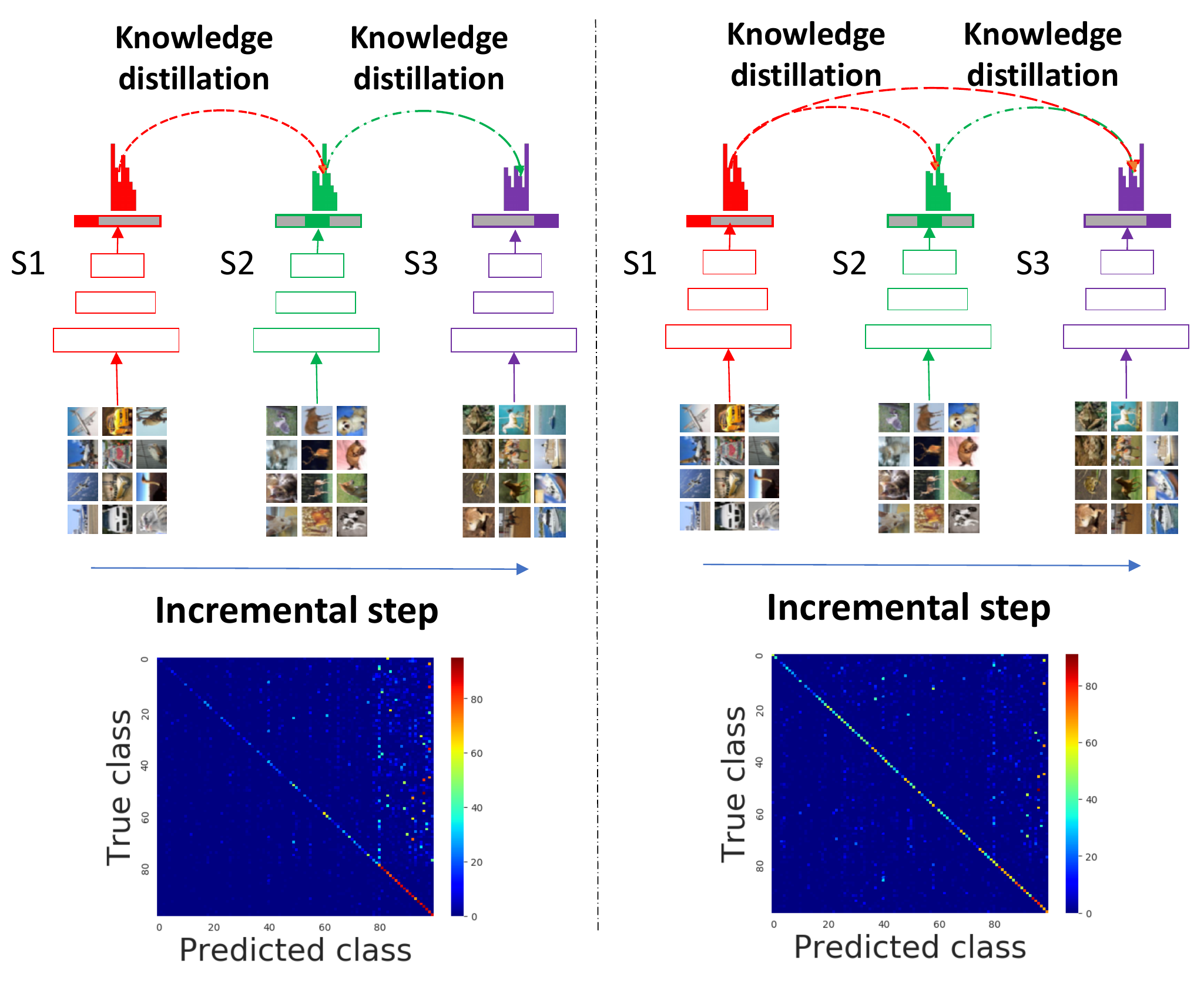}
\end{center}
 \vspace{-10pt}
   \caption{Concept overview. We propose to distill knowledge from all previous models efficiently to preserve old data information rather than sequentially applying distillation only to the last model. (\eg. using both S1 and S2 in S3 for distillation instead of sequentially using S1 for S2 and then S2 for S3). The confusion matrix is LWF-MC \cite{rebuffi2017icarl} on the left and our method on the right for the exemplar-free incremental setting.}
\label{fig:overview}
\vspace{-12pt}
\end{figure}

Prior methods typically apply knowledge distillation~\cite{hinton2015distilling} sequentially during the incremental procedure to preserve previous knowledge. Since they apply distillation only to the penultimate model, it is difficult to maintain all past knowledge completely (the left side of Figure~\ref{fig:overview}). From that observation, we propose using all the model snapshots. Prior knowledge is preserved better through our approach (the right side of Figure~\ref{fig:overview}). However, saving all previous models may incur a great penalty in memory storage and without somehow compressing this historical information would not be practical. To address this, we reconstruct previous outputs using only ``necessary" parameters during training.

\begin{figure*}[t]
\centering
   \includegraphics[width=0.8\linewidth]{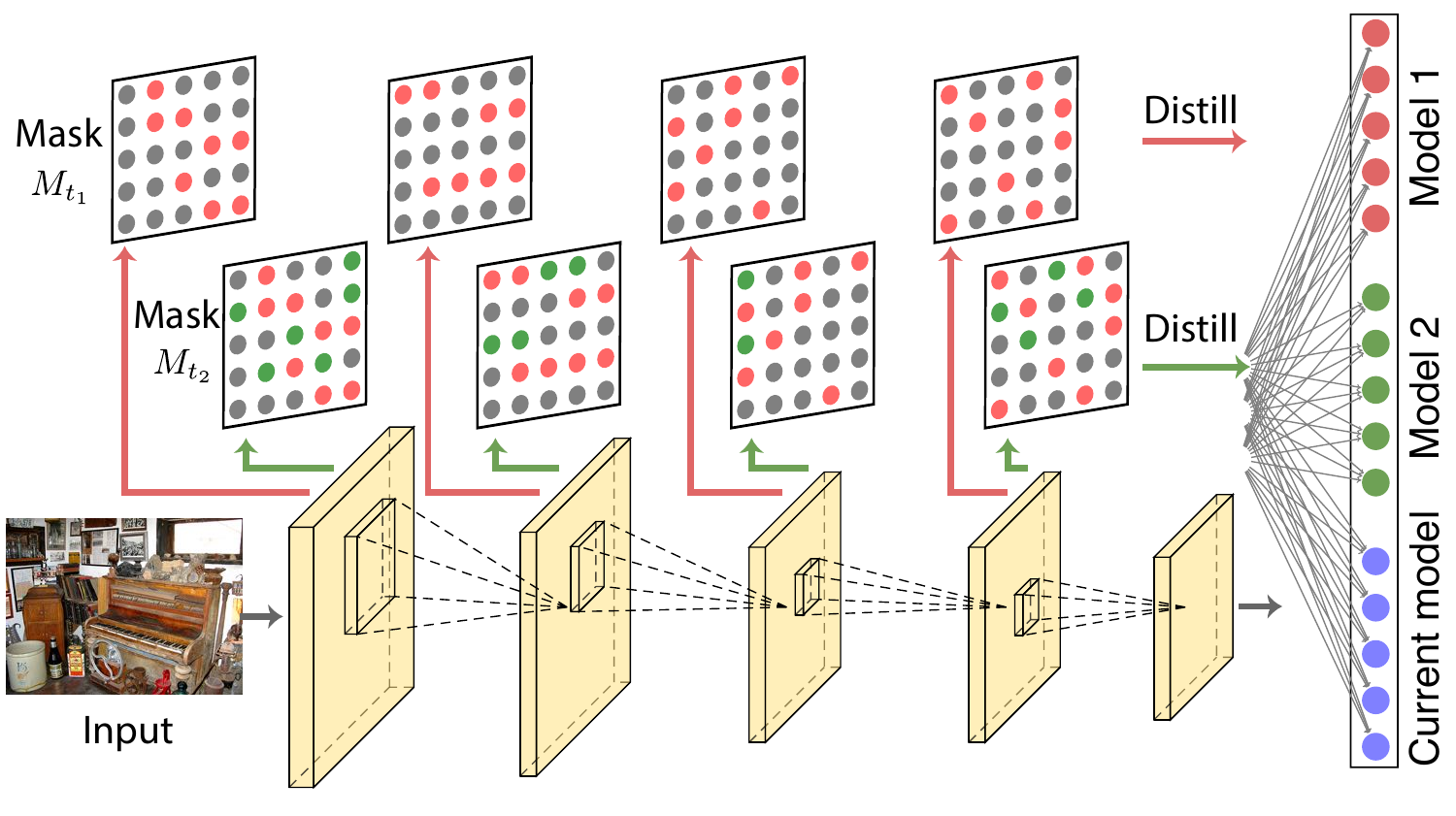}

   \caption{Framework overview. Given images from the current training data, we preserve previous knowledge directly from the reconstructed output through matching the logits with the corresponding model (\eg, model 1 and 2) and classifying the current data with its ground truth. The gray dots represent the weights to be trained on the current data. The red and green dots are fixed during training, denoting the weights retained from the first and second incremental step respectively.}
\label{fig:frame}
\vspace{-15pt}
\end{figure*}

To this end, we propose an end-to-end Multi-model and Multi-level Knowledge Distillation ($\boldsymbol M^2$KD) framework as depicted in Figure~\ref{fig:frame}. We introduce a multi-model distillation loss which leverages the snapshots of all previous models to serve as teacher models during distillation, and then directly matches the outputs of a network with those from the corresponding teacher models. To make the pipeline more efficient, we adapt mask based pruning methods to reconstruct the previous models. We prune the network after each incremental training step and identify significant weights to reconstruct the model. This allows us to reconstruct previous models on-the-fly and utilize them as teacher models in our multi-model distillation. To further enhance the distillation process, we also include an auxiliary distillation loss to preserve more intermediate features of previous models. Additionally, our approach addresses catastrophic forgetting in sequential distillation, and thus generalizes well for both exemplar based and exemplar-free settings.

To show the effectiveness of our approach, we evaluate our model on Cifar-100~\cite{krizhevsky2009cifar} and a subset of ImageNet~\cite{krizhevsky2012imagenet}. We achieve state-of-the-art performance for all the datasets in exemplar-free setting. We also show improvement when adapting to exemplar-based incremental learning and our exemplar-free setting outperforms iCaRL~\cite{rebuffi2017icarl} with a 200 exemplar budget. 

In summary, our contributions are three fold. First, we propose a multi-model distillation loss, which directly matches logits of the current model with those from the corresponding teacher models. Secondly, for efficiency, we reconstruct historical models via mask based pruning such that model snapshots can be reconstructed with low memory footprint. Experiments on standard incremental learning benchmarks show that our method achieves state-of-the-art performance in exemplar-free incremental setting.


\section{Related Work}
The ultimate goal of incremental learning is to achieve good performance on new data while preserving the knowledge about old data. Generally, two types of evaluation settings~\cite{chaudhry2018riemannian} have been considered. One is multi-head incremental learning which utilizes multiple classifiers at inference, and the other is single-head incremental learning which only utilizes one classifier at inference. 

\noindent\textbf{Multi-head incremental learning.} The evaluation setting in this stream is that a specific classifier is selected during testing according to the tasks or categories. With this prior information, no confusion exists across different classifiers, and thus the target becomes how to adapt the old model for new tasks or categories. Research has been focused on utilizing an episodic memory to trace back previous tasks~\cite{lopez2017gem,AGEM,aljundi2018mas}, or constraining the important weights on old tasks~\cite{kirkpatrick2017ewc}. In addition, Mallya \etal~\cite{mallya2018packnet,mallya2018piggyback} learn a mask for pruning to further constrain the weights on old tasks. Hou \etal~\cite{hou2018retrospection} distill the knowledge from the old model when adapting to new tasks. Different from this setting, we do not assume the task or category information is known during inference and follow the setting of single-head incremental learning. Even though we apply pruning in our approach, our goal is different from~\cite{mallya2018packnet,mallya2018piggyback} as the masks are utilized to reconstruct previous models and our approach requires no mask selection at inference. 

\noindent\textbf{Single-head incremental learning.} Single-head evaluation uses only one classifier to predict both the old and the new classes. This setting is more challenging \cite{chaudhry2018riemannian} compared to the multi-head counterpart because of the confusion between old and new categories. Knowledge distillation \cite{hinton2015distilling} is frequently utilized to preserve information. Li \etal~\cite{li2018lwf} distill the knowledge from the penultimate model. Dhar \etal~\cite{dhar2018lwm} introduce Grad-CAM \cite{selvaraju2017grad} in the loss function. A relaxed setting is to introduce exemplar set~\cite{rebuffi2017icarl} for the old data and match previous logits through distillation. Castro \etal~\cite{Castro2018end2end} explore the balance between old and new data during training. Li \etal~\cite{li2018supportnet} focus on constructing exemplar set and Caselles \etal~\cite{caselles2018replay,ostapenko2019ltr} replay the seen categories with GANs \cite{goodfellow2014generative}. 
Instead of saving exemplars, we save the parameters of previous models for reconstruction. With that, this paper can be considered a complement research direction. In fact, as knowledge distillation is an important component in these methods, they can potentially benefit from our approach as well. Additionally, Javed \etal~\cite{javed2018revisiting} alleviate the bias in knowledge distillation by introducing a scaling vector to trained classifier, however, our approach is agnostic to classifier and achieves better performance.

\noindent\textbf{Network pruning.} Considerable research has explored this area to reduce network redundancy. Han \etal~\cite{han2016dsd,han2015weights} propose to compress network through quantization and Huffman coding. Yu \etal~\cite{yu2018nisp} compress the weights according to their scores. Other methods~\cite{molchanov2016pruning,jaderberg2014speeding,liu2017inference} explore compression for fast inference. In contrast to these methods, we leverage network redundancy and use pruning to reconstruct all previous models in incremental learning with low memory footprint.

\section{Approach}
We propose novel distillation losses to preserve previous information without introducing too much memory overhead (See Figure~\ref{fig:frame}). The model is agnostic to the backbone architecture and generalizes well to both exemplar based and exemplar-free methods.

\subsection{Multi-model Distillation}

Single-head incremental learning consists of a sequence of incremental class inclusion process, referred to as incremental steps. Samples from a batch of new classes $C_k$ are added at the $k$-th incremental step. For instance, 20 classes will be added per incremental step in a 20-class batch setting. Accordingly, the network assigns new logits (output nodes) for the incremental classes. At inference, the maximum logit score in the output is treated as the final decision.

The knowledge distillation used in incremental learning \cite{li2018lwf,rebuffi2017icarl} mainly aims to match the output of the current model to a concatenation of the penultimate model logits and ground truth labels. Formally, it optimizes the cross entropy for both the old and new logits,

\begin{align}
    L_{D} & =-\frac{1}{N}\sum_{i=1}^{N}\sum_{j=1}^{C_o} s^{'}_{ij}\log(s_{ij})
    \nonumber \\
    & - \frac{1}{N}\sum_{i=1}^{N}\sum_{j=C_o + 1}^{C} y_{ij}\log(s_{ij}),
\end{align}
where $N$ and $C$ denotes the number of samples and the total class number so far respectively, and $C_o$ denotes the old classes. $s_{ij}$ is the output score of the network obtained by applying Sigmoid function to the output logits for sample $i$ at logit $j$. $s^{'}_{ij}$ denotes the old score obtained by the penultimate model. $y_{ij}$ denotes the ground truth. 

Treating the penultimate model as the teacher and applying this distillation sequentially helps preserve historical information, especially when no previous exemplar set is stored, which is the protocol for prior methods \cite{rebuffi2017icarl,Castro2018end2end,li2018lwf,dhar2018lwm}. However, the historical information will be gradually lost in this sequential pipeline as the current model must reconstruct all the prior information from the penultimate model alone. To address this limitation, we propose multi-model distillation, which directly leverages all previous models as our teacher model set. Since we mainly have current training data and labels for both settings, the network is more confident on current classes than old ones. Therefore, matching the previous logits of the current model directly with their corresponding old models preserves information better than always using the penultimate model. Formally, we minimize the cross entropy for the logits between the current model and corresponding teacher models from previous incremental steps,
\begin{align}
    L_{MMD} & =-\frac{1}{N}\sum_{i=1}^{N}\sum_{k=1}^{P-1}\sum_{j=C_{k-1}+1}^{C_k} s^{'}_{ijk}\log(s_{ijk})
    \nonumber \\
    & - \frac{1}{N}\sum_{i=1}^{N}\sum_{j=C_{P-1} + 1}^{C} y_{ij}\log(s_{ij}),
\end{align}

where classes from $C_{k-1}+1$ to $C_k$ belong to the $k$-th incremental step and $P$ denotes the number of incremental steps. Classes from $C_{P-1}+1$ to $C$ belong to the current categories. $s_{ijk}$ is the output score of the current model for sample $i$ at logit $j$ in the $k$-th incremental step. $s^{'}_{ijk}$ denotes the output score of the $k$-th previous model. 

At inference, we directly choose the maximum among the output logits, which acts as an ensemble of all the previous teacher models and the current model.

\begin{figure}[t]
\centering
   \includegraphics[width=0.7\linewidth]{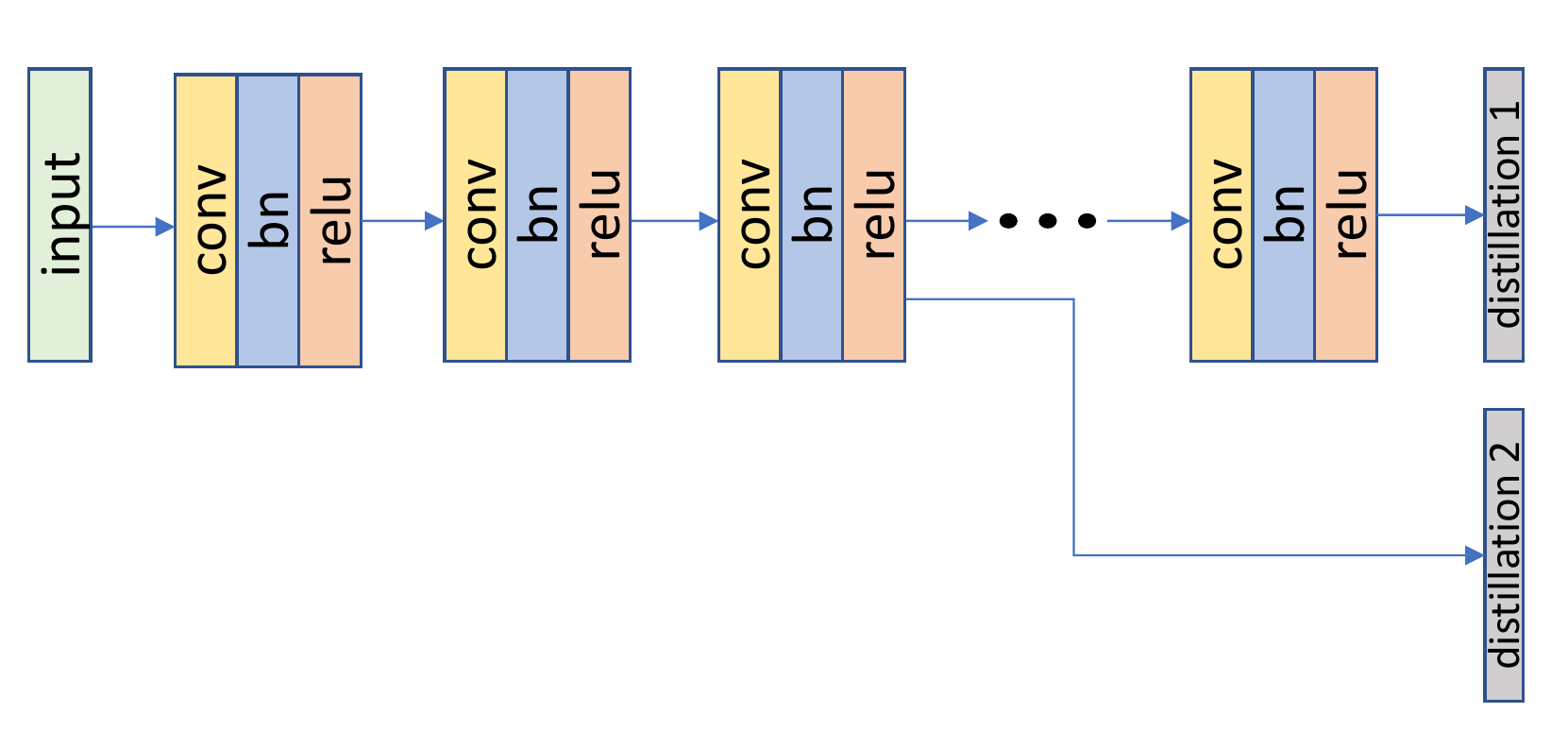}

   \caption{Illustration of auxiliary distillation. We extract the intermediate features and connect directly with an auxiliary classifier to preserve middle level knowledge. }

\label{fig:auxiliary}
\end{figure}

\subsection{Auxiliary Distillation} 

Previous incremental learning methods preserve old class information through matching the final output. However, the features from intermediate layers also contain useful information. Inspired by the auxiliary loss in segmentation task~\cite{zhao2017pyramid}, we propose an auxiliary distillation loss to preserve the intermediate statistics of previous models. Similar to using the final output to represent network statistics, the prediction made by lower level features also represents intermediate feature statistics. Following the main branch classification, we extract lower level features and use an auxiliary classifier to conduct classification based on intermediate features (See Figure~\ref{fig:auxiliary}). 

Also, a multi-model distillation loss is added to this auxiliary classifier for the purpose of preserving prior lower level features, and a standard cross entropy loss is included for classifying the current data. Formally,

\begin{align}
    L_{AD} & =-\frac{1}{N}\sum_{i=1}^{N}\sum_{k=1}^{P-1}\sum_{j=C_{k-1}+1}^{C_k} a^{'}_{ijk}\log(a_{ijk})
    \nonumber \\
    & - \frac{\alpha}{N}\sum_{i=1}^{N}\sum_{j=1}^{C} y_{ij}\log(a_{ij}),
\end{align}

where $a^{'}_{ijk}$ denotes the output score from previous auxiliary classifiers, $a_{ijk}$ or $a_{ij}$ is the output score of the auxiliary branch, $\alpha$ is the ratio between the distillation and cross entropy loss. Notice that all the logits in ground truth labels are utilized in the classification cross entropy to enforce the correct prediction of current data.

The total loss function of the network becomes,
\begin{equation}
    L_{total}= L_{MMD} + \lambda L_{AD},
\end{equation}
where $\lambda$ is the ratio between the main classification multi-model distillation and the auxiliary classification distillation. This auxiliary classification branch is only used during training. At inference time, we only use the main branch classifier for prediction.

\subsection{Model Reconstruction}

One drawback of multi-model distillation in its original form is that it utilizes all previous models, requiring additional memory storage for the models. However, we observe that distillation aims to match logits. Therefore it is only necessary to preserve the outputs of previous networks, not the entire networks themselves. Our idea is to save only a small set of the necessary parameters from which we can approximate the output. By that way, all the models can be recovered on-the-fly without large memory penalty. 

To determine the necessary parameters, we adapt mask based pruning~\cite{mallya2018packnet} for model reconstruction. Specifically, after training each incremental step we sort the magnitude of weights in each layer, freeze the important ones to reach a specified pruning ratio, and use the residual weights to train the next incremental class set. We repeat this procedure for all future incremental steps until all the incremental classes are included.

Formally, the output of a network with $n$ convolutional layers is obtained from its classifier (the last layer) and features,

\begin{equation}
\label{eq:ld}
    s = \Psi(f^{(n)}),
\end{equation}
where $\Psi$ denotes the classifier and $f^{(n)}$ denotes the features in the $n$-th layer and can be generally written as
\begin{equation}
\label{eq:ld}
f^{(n)}=\sigma(w^{(n)}f^{(n-1)}+b^{(n)}),
\end{equation}
where $w$ and $b$ are weights and biases respectively, $\sigma$ denotes the activation function and $f^{(0)}$ is the input.

We use a mask $M$ to identify the important weights of each layer for all previous incremental steps. After each pruning procedure, we update the mask for the current incremental step, \eg, using $k$ to mark the weights for the $k$-th step. With the mask $M_{k}$ for the $k$-th incremental step, we reconstruct the corresponding features by:

\begin{equation}
\label{eq:ld}
    f_{k}^{(n)}=\sigma(w_{k}^{(n)}\delta(M_{k}^{(n)}<=k)f_{k}^{(n-1)}+b_{k}^{(n)}),
\end{equation}
where $M_{k}^{(n)}$ denotes the mask in the $n$-th layer at incremental step $k$, $f_{k}^{(n)}$ denotes the feature in the $n$-th layer in $k$-th incremental step, and $\delta$ denotes delta function.

With the saved biases, batch normalization and classifier parameters, we can reconstruct all previous models from the pre-updated model on-the-fly. The output of the $k$-th model is reconstructed by

\begin{equation}
\label{eq:ld}
    s_k = \Psi_k(f_{k}^{(n)}),
\end{equation}
where $s_k$ and $\Psi_k$ denote the output of the network and the classifier for the $k$-th incremental step respectively.


\section{Experiments}\label{experiment}
We first evaluate our method in the exemplar-free setting. Then we extend our method to the exemplar-based setting. For more analysis, we also compare our memory cost with other methods.

\subsection{Implementation Details}
We use PyTorch for implementation. The network architecture follows prior works \cite{rebuffi2017icarl,li2018lwf,Castro2018end2end}: we use ResNet-32 \cite{he2016deep} with input size $32 \times 32$ for Cifar-100 and ResNet-18 with input size $224 \times 224$ for iILSVRC-small. We extract the output of the second residual block for auxiliary distillation and empirically set $\alpha$ to 0.5. We train 80 epochs for Cifar-100 and 60 epochs for iILSVRC-small. Following the setting in \cite{rebuffi2017icarl}, we use the training batch size of 128 and the initial learning rate of 2.0 to train the model. The learning rate decays by a factor of 5 every 40 epochs for Cifar-100 and 20 epochs for iILSVRC-small. Weight decay with a factor of 1e-5 is applied for the first incremental step and 0 for the rest in our full model to ensure weights from previous models remain the same. We optimize the network using standard Stochastic Gradient Descent (SGD) with a momentum 0.9. The pruning ratio is 0.75 for class batch less than 20 and 0.8 for 20 groups. After cutting off insignificant weights in the pruning stage, we fine tune the network for another 15 epochs. $\lambda$ is set to be 1.0 to balance the losses. Only random horizontal flipping is applied as data augmentation for all experiments.

\subsection{Datasets and Evaluation Metrics}

The evaluation is conducted on iILSVRC-small~\cite{russakovsky2015imagenet} and Cifar-100~\cite{krizhevsky2009cifar}. 

\noindent\textbf{Evaluation Metrics.}
Following the same metrics in prior methods~\cite{li2018lwf,rebuffi2017icarl}, the top-1 classification accuracy is reported for Cifar-100 and top-5 classification accuracy is reported for iILSVRC-small.

 \begin{figure*}[t]
\centering
   \begin{subfigure}[Cifar-100 5-class batch ]{\centering\includegraphics[width=110pt]{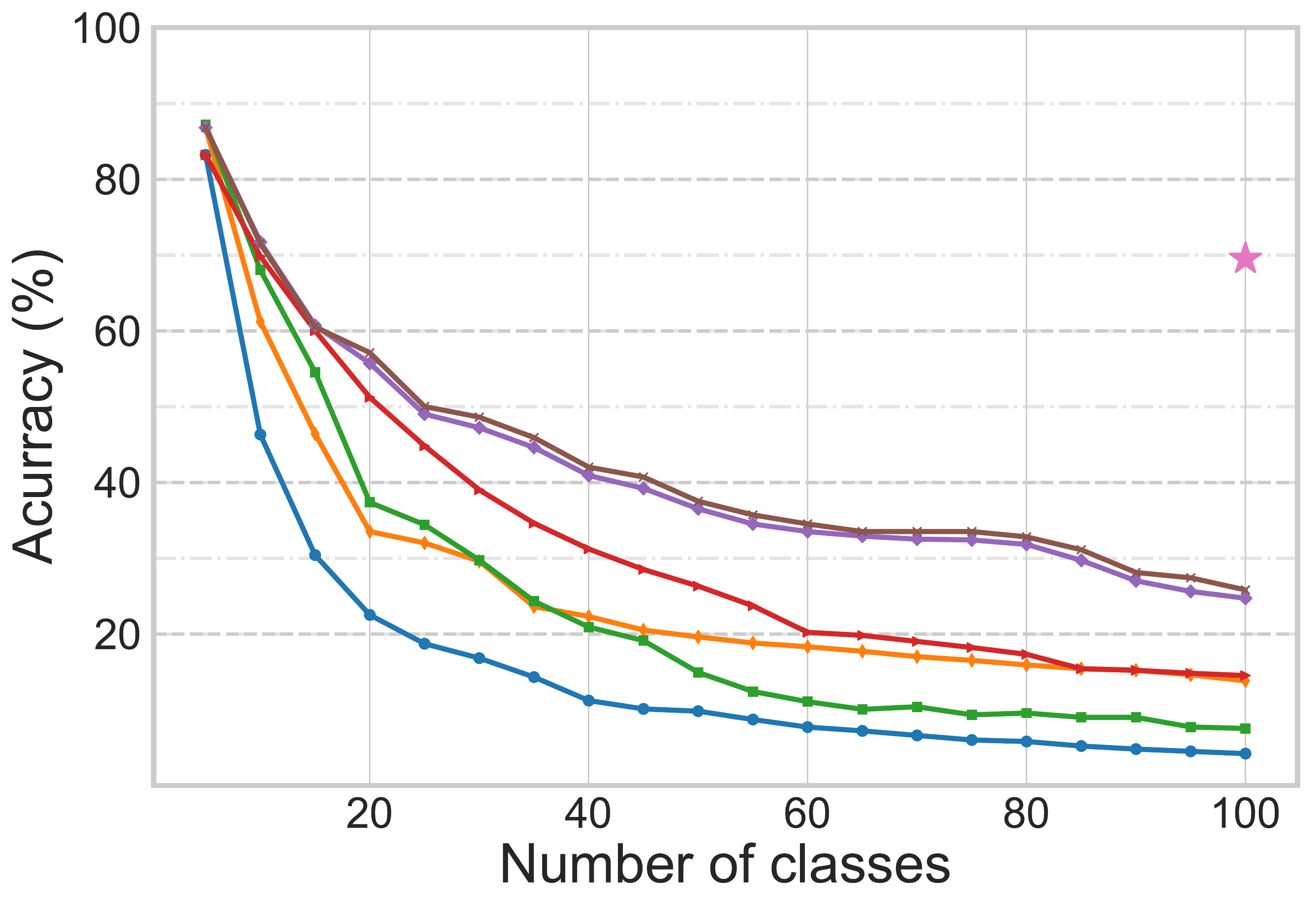}}
   \end{subfigure}   
   \begin{subfigure}[Cifar-100 10-class batch ]{\centering\includegraphics[width=110pt]{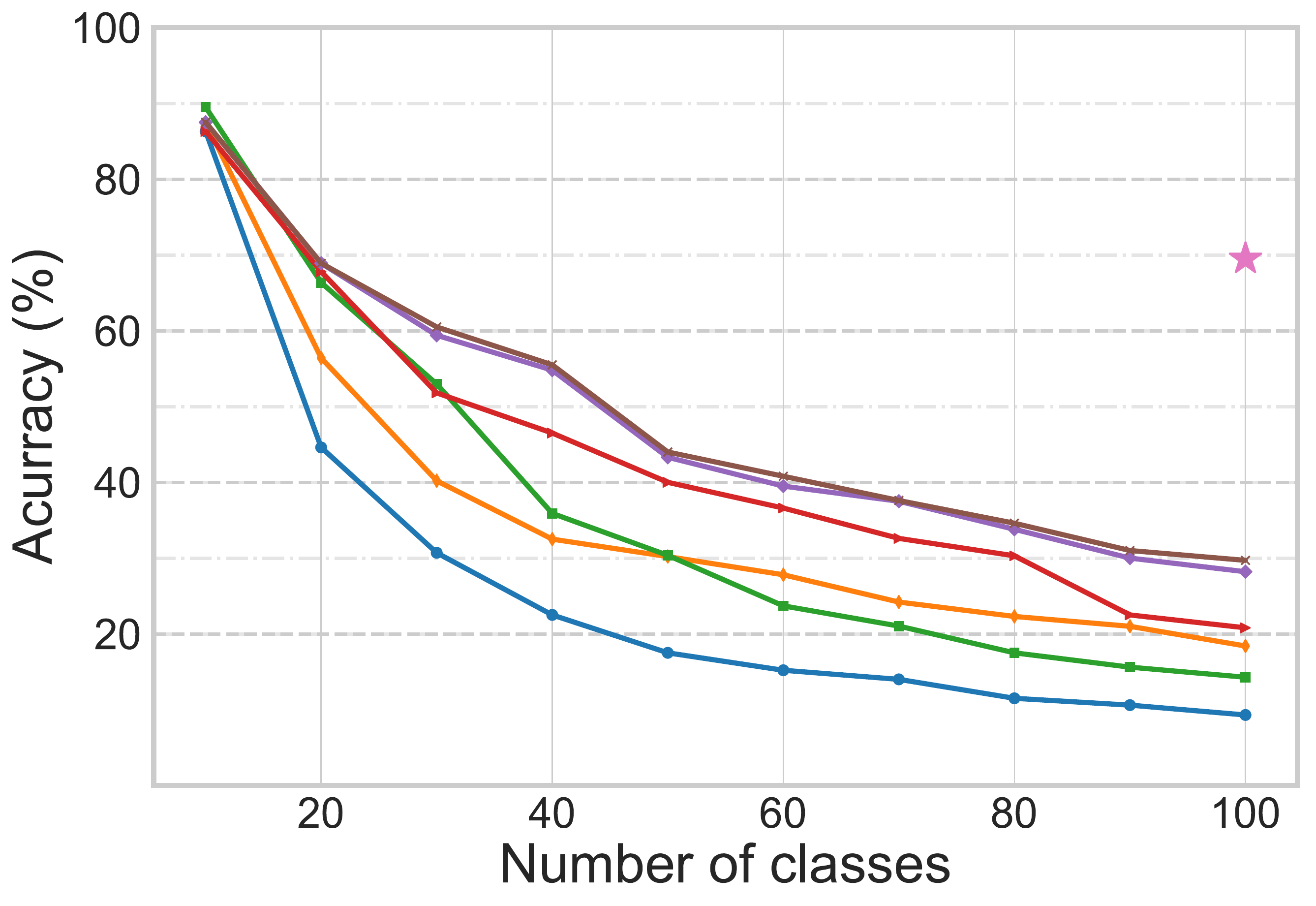}}
   \end{subfigure} 
   \begin{subfigure}[Cifar-100 20-class batch ]{\centering\includegraphics[width=110pt]{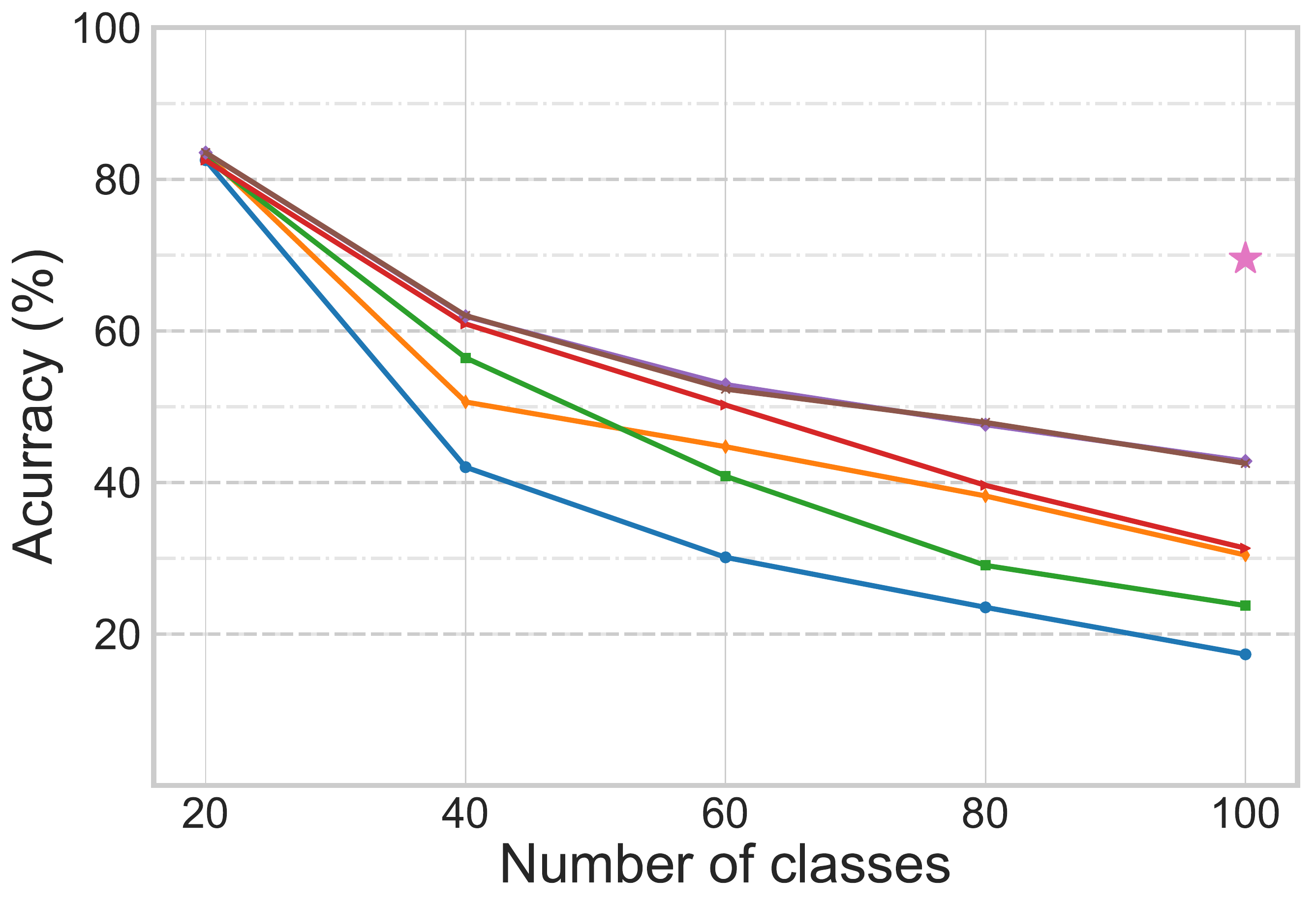}}
   \end{subfigure} 
   \begin{subfigure}[iILSVRC-small 10-class batch ]{\centering\includegraphics[width=110pt]{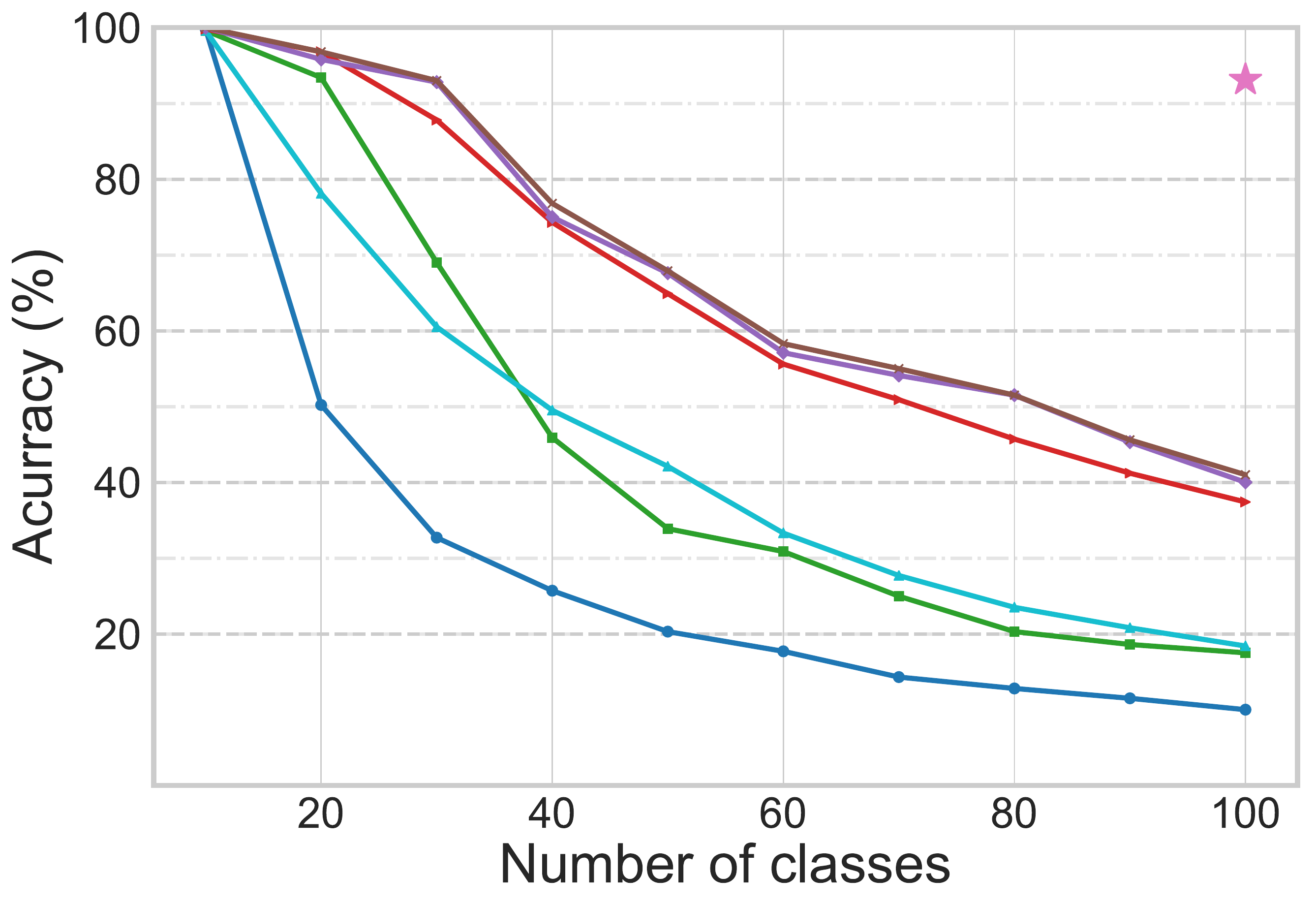}}
   \end{subfigure} 
   \begin{subfigure}[iILSVRC-small 20-class batch ]{\centering\includegraphics[width=110pt]{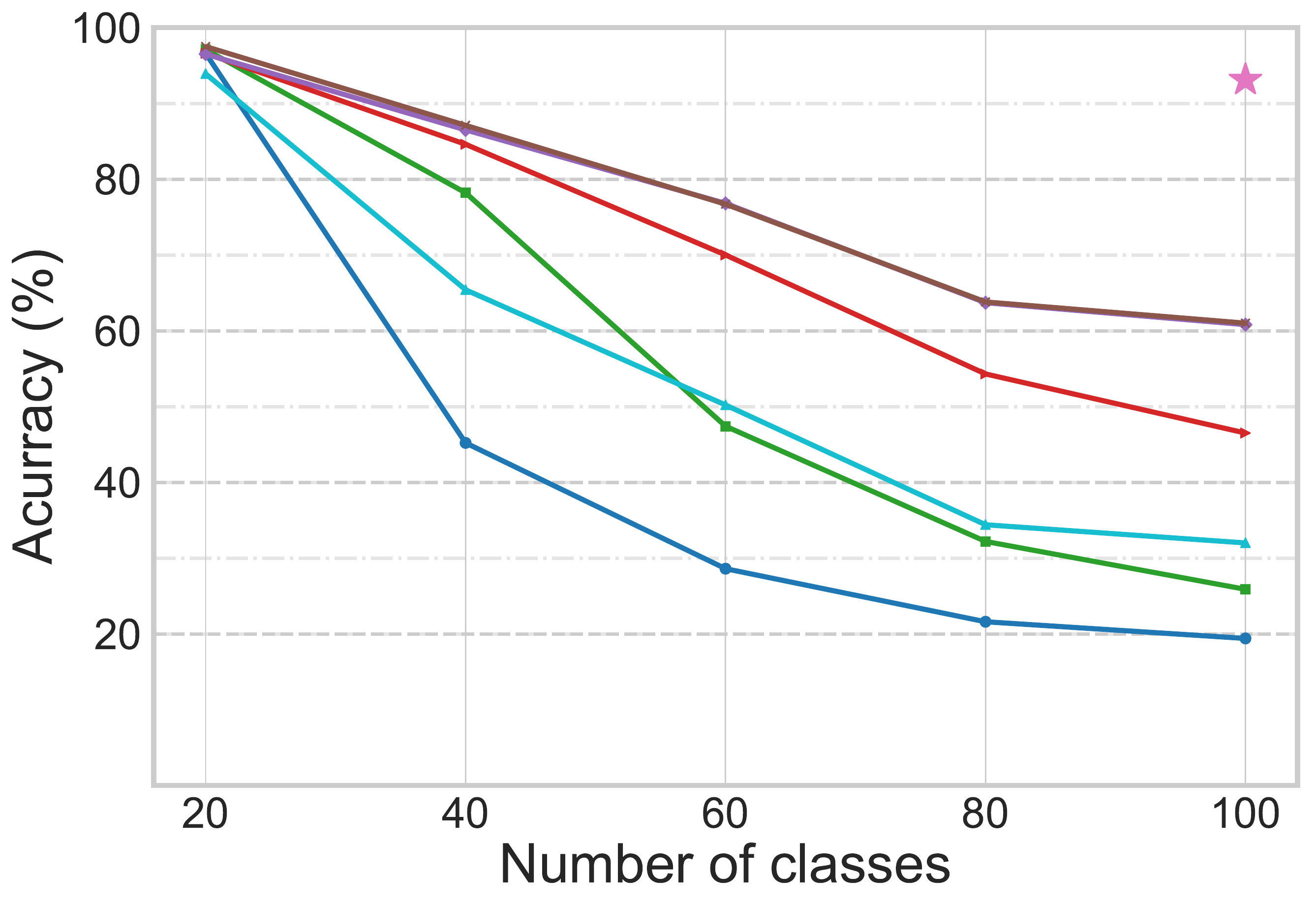}}
   \end{subfigure} 
   \begin{subfigure}[Legend]{\centering\includegraphics[width=80pt]{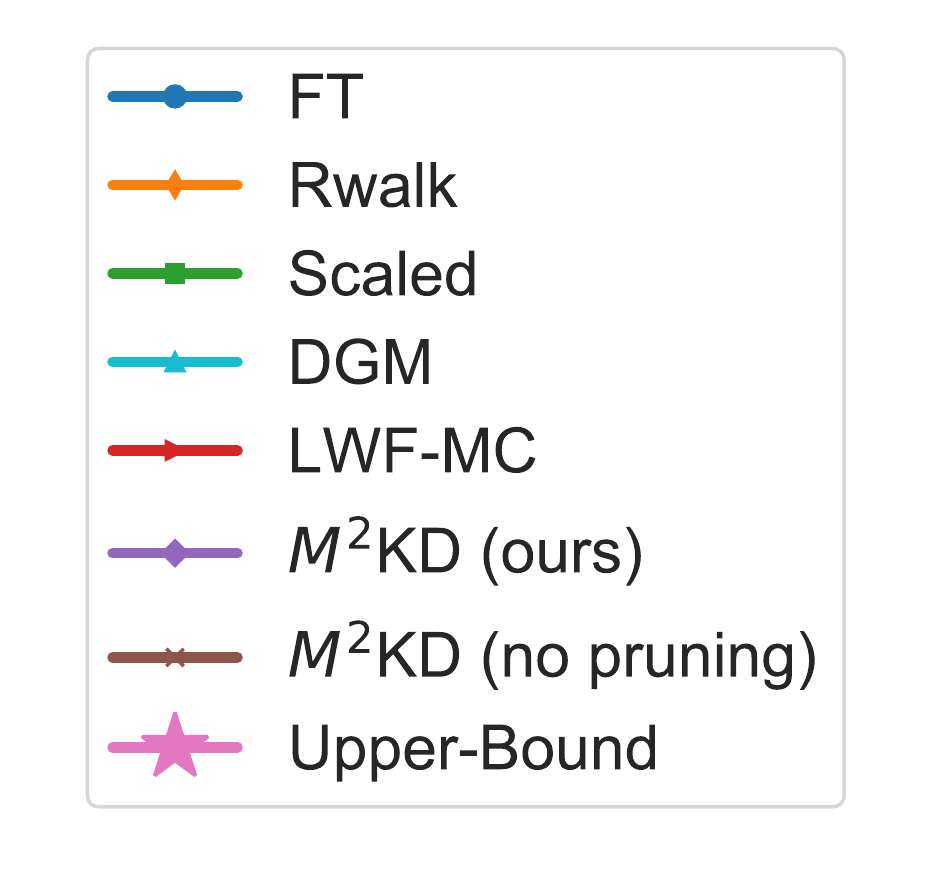}}
   \end{subfigure} 
   \caption{Performance on iILSVRC-small and Cifar-100 dataset in exemplar-free setting.  \textbf{(a)} Top-1 accuracy on Cifar-100 (5-class batch).  \textbf{(b)} Top-1 accuracy on Cifar-100 (10-class batch).  \textbf{(c)} Top-1 accuracy on Cifar-100 (20-class batch).  \textbf{(d)} Top-5 accuracy on iILSVRC-small (10-class batch). \textbf{(e)} Top-5 accuracy on iILSVRC-small (20-class batch). }

\label{fig:main}
\vspace{-15pt}
\end{figure*}

\subsection{Exemplar-free setting}\label{noexemplar}
We evaluate our methods in exemplar-free single-head setting. For evaluation, we also compare with state-of-the-art single-head approaches--- \textbf{Scaled}~\cite{javed2018revisiting},\textbf{DGM}~\cite{ostapenko2019ltr},\textbf{Rwalk}~\cite{chaudhry2018riemannian},\textbf{LWF-MC}~\cite{rebuffi2017icarl} --- and the following baselines. (DGM hasn't released code for Cifar-100 and thus is not compared in the Cifar-100 experiment.)

\noindent\textbf{FT}: A baseline approach that only applies cross entropy loss to fine-tune the penultimate model on new coming incremental classes. Knowledge distillation is not applied. 

\noindent\textbf{$\boldsymbol M^2$KD (ours)}: Our full model applying multi-model, auxiliary distillation along with pruning to save memory storage.

\noindent\textbf{$\boldsymbol M^2$KD (no pruning)}: Our full model which directly loads all the previous snapshots for multi-model distillation.

\noindent\textbf{Upper-Bound}: The upper bound which directly trains all classes together. 

Figure~\ref{fig:main} highlights our performance compared to state-of-the-art methods. For Cifar-100, our method consistently outperforms other methods from 5-class to 20-class batch per incremental step. The margin becomes larger as more incremental steps are added. This demonstrates the advantage of multi-model distillation as it avoids accumulating loss of historical information. Similar observation can be made when evaluating on iILSVRC-small. It is interesting to note that our model with pruning achieves comparable performance with the no-pruning version. This indicates the effectiveness of the pruning procedure in terms of saving memory while maintaining performance. Even though the residual active weights decrease gradually due to pruning, we still preserve the performance up to 20 incremental steps. 

\subsection{Analysis on pruning ratio}\label{pruning}
We compare the results corresponding to different pruning ratios to investigate the robustness of our approach. Table~\ref{tab:prune} summarizes the results. Marginal performance variation (around $3\%$) is observed for different pruning ratios. Even though a higher (0.9) pruning ratio affects the performance as the active weights decrease in the current incremental step and a lower (0.6) ratio affects the performance as available weights decrease in the future steps, the relatively trivial influence indicates that a large redundancy exists in the network architecture. Benefitting from it, our approach shows robustness to different pruning ratios.

\begin{table}[t]
\begin{minipage}{0.55\textwidth}
\centering 
\small
\begin{tabular}{p{1.5cm}lccccc} 
\hline
 Step &  1 &  2 &  3 &  4 & 5 \\ \hline
No pruning & \textbf{83.5} & \textbf{61.8} &\textbf{52.5} &\textbf{51.5} & 42.1\\
 Ratio 0.6 & 82.9 & 59.6 &52.2 &46.5 & 40.1\\
 Ratio 0.7 & 83.5 & 61.7&52.5 &50.0 & \textbf{42.8} \\
 Ratio 0.8 & 83.5 &58.5 & 52.0& 49.3 &42.0 \\
 Ratio 0.9 & 83.0 & 58.0 & 49.7 & 47.3 & 39.9\\\hline 
\end{tabular}

\caption{Top-1 accuracy comparison among different pruning ratios on Cifar-100 (20 classes per incremental step). }

\label{tab:prune}
\end{minipage}\hfill
\begin{minipage}{0.45\textwidth}
\centering  
\small
\begin{tabular}{p{1.3cm}lcc} 
\hline
  & iILSVRC-small & Cifar-100 \\ \hline
 LWF-MC &0 &0\\
 iCaRL  &68.0 &9.4 \\
Ours &9.80 &0.84 \\\hline 
\end{tabular}
\caption{Memory compensation comparison (MB). Each entry is the additional memory requirement for methods across different datasets based on the memory footprint of LWF.}
\label{tab:storage}
\end{minipage}
\vspace{-15pt}
\end{table}

\subsection{Exemplar Based Setting}\label{exemplar}
Our approach can also be applied to exemplar based incremental learning methods which use distillation sequentially on the output of networks~\cite{rebuffi2017icarl,Castro2018end2end,li2018supportnet}. To evaluate our model in this setting, we add exemplar selection to our approach and compare with exemplar based methods. 

\noindent\textbf{iCaRL}~\cite{rebuffi2017icarl}: A prominent exemplar based incremental learning approach which constructs exemplar set for the old data according to the feature means and do distillation on the penultimate model. A nearest class mean classifier \cite{mensink2012ncm} is applied at inference. 

\noindent\textbf{iCaRL aux}: Adding auxiliary distillation to  iCaRL.

\noindent\textbf{iCaRL $\boldsymbol M^2$KD}: Change the original distillation function which only matches logits from the penultimate model to our multi-model distillation. Auxiliary distillation is also appended for a better performance.

\begin{figure}[]
\centering

   \begin{subfigure}[Top-1 Cifar-100]{\centering\includegraphics[height=100pt,width=120pt]{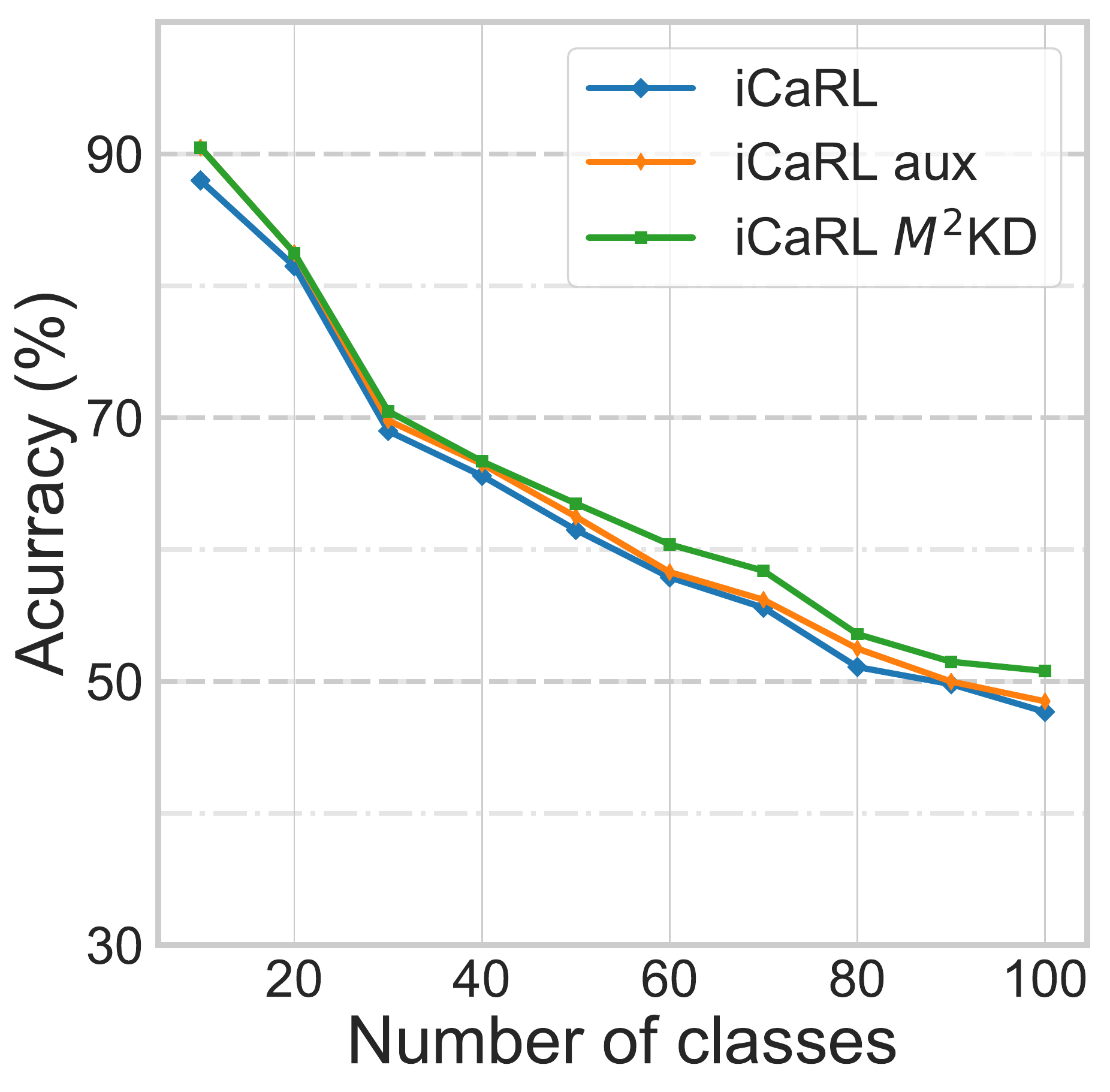}}
   \end{subfigure}
    \begin{subfigure}[Top-5 iILSVRC-small]{\centering\includegraphics[height=100pt,width=120pt]{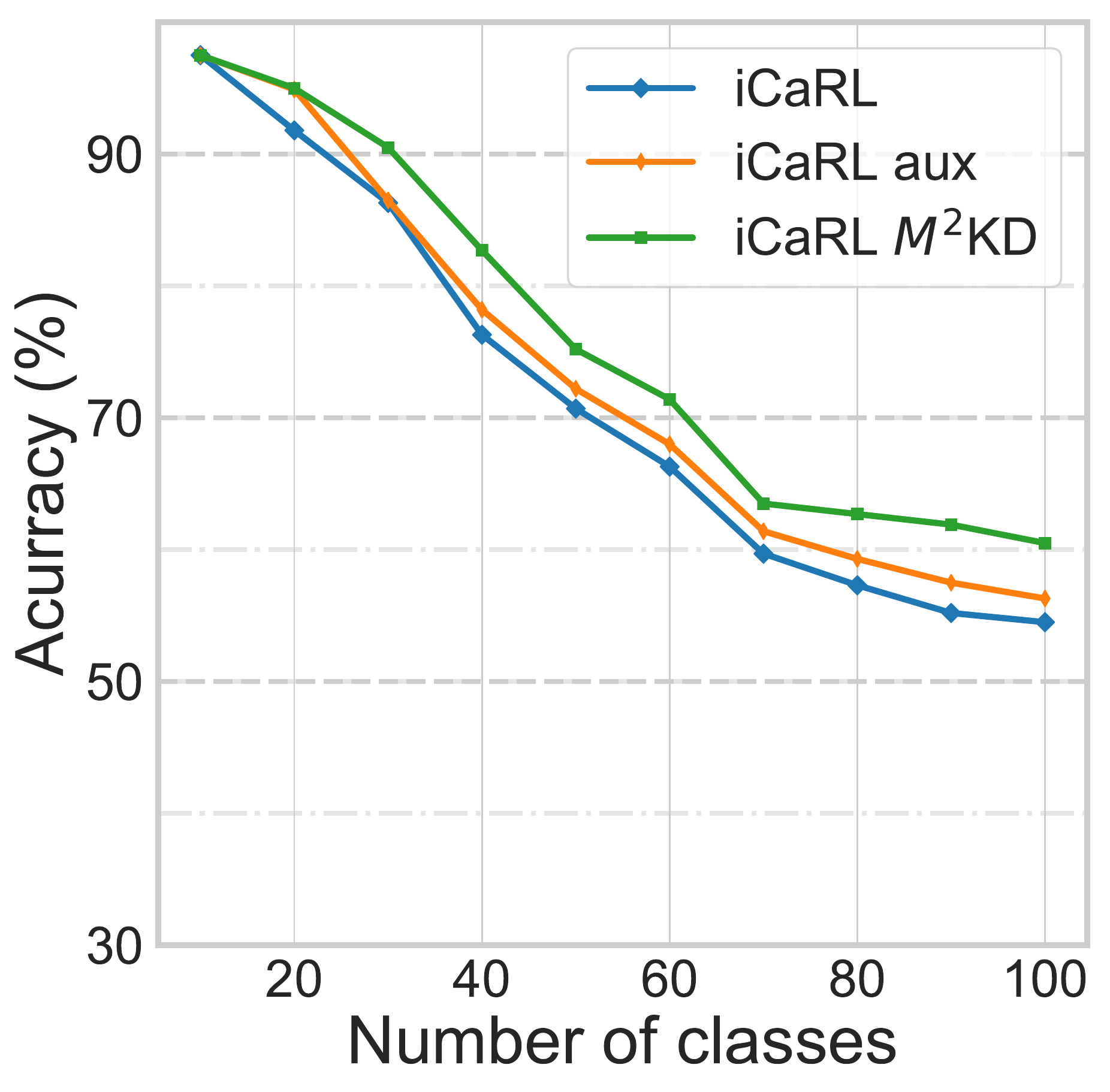}}  
    \end{subfigure} 
   \caption{Performance comparison in exemplar based setting (10-class batch).  \textbf{(a)} Top-1 accuracy performance on Cifar-100.  \textbf{(b)} Top-5 accuracy performance on iILSVRC-small. }
  
\label{fig:icarl}
\vspace{-15pt}
\end{figure}

The results are shown in Figure~\ref{fig:icarl}. With the introduction of multi-model and auxiliary distillation, the performance of iCaRL improves. It indicates that with direct access to all the previous models for distillation, the knowledge preserves better even with exemplar set. 

\subsection{Memory Comparison}\label{memory}

Starting from the memory footprint of LWF as our baseline, we compare the extra memory storage between exemplar based method such as iCaRL~\cite{rebuffi2017icarl} and our approach. The memory is calculated in the 10-class incremental step setting for both iILSVRC-small and Cifar-100. For our approach, we directly calculate the storage difference between the last and the initial step. For iCaRL, the memory is approximately calculated by the average size of image for 2000 samples (\textit{i.e.} the default exemplar size), and the compensation for saving the record of exemplar set. To optimize the memory consumption of iCaRL, we resize the images in iILSVRC-small to $256 \times 256$ and compress to JPG with quality 95 to match their network input size during training.

Table~\ref{tab:storage} shows the memory compensation for different methods. It indicates that our approach has approximately $7 \times$ smaller memory compensation on iILSVRC-small and $10 \times$ smaller on Cifar-100 than iCaRL. On average, for each incremental step, our approach only takes 0.98 MB and 0.08 MB for iILSVRC-small and Cifar-100 respectively. The memory advantage to exemplar based methods might become larger as higher resolution images take more storage.


\section{Conclusion and Discussion}
This paper presents a novel distillation strategy that mitigates catastrophic forgetting in single-head incremental learning setting. We introduce multi-model distillation which directly guides the model to distill knowledge from the corresponding teacher models. To further improve our performance, we incorporate auxiliary distillation to preserve intermediate features. More efficiently, we avoid saving all the model snapshots through reconstructing all previous models using mask based pruning algorithm. Extensive experiments on standard incremental learning benchmarks demonstrate the effectiveness of our approach. Incremental learning is still far from solved. A significant gap between one-step training versus incremental training still exists. It remains to be a open question how to reduce the confusion between different incremental steps especially without access to previous data, which might be a future exploration for our research.

\noindent\textbf{Acknowledgement}
This work was partly funded by Adobe. The authors
acknowledge the Maryland Advanced Research Computing
Center (MARCC) for providing computing resources.

\bibliography{egbib}

\begin{thebibliography}{37}
\providecommand{\natexlab}[1]{#1}
\providecommand{\url}[1]{\texttt{#1}}
\expandafter\ifx\csname urlstyle\endcsname\relax
  \providecommand{\doi}[1]{doi: #1}\else
  \providecommand{\doi}{doi: \begingroup \urlstyle{rm}\Url}\fi

\bibitem[Aljundi et~al.(2018)Aljundi, Babiloni, Elhoseiny, Rohrbach, and
  Tuytelaars]{aljundi2018mas}
Rahaf Aljundi, Francesca Babiloni, Mohamed Elhoseiny, Marcus Rohrbach, and
  Tinne Tuytelaars.
\newblock Memory aware synapses: Learning what (not) to forget.
\newblock In \emph{ECCV}, 2018.

\bibitem[Caselles-Dupr{\'e} et~al.(2018)Caselles-Dupr{\'e}, Garcia-Ortiz, and
  Filliat]{caselles2018replay}
Hugo Caselles-Dupr{\'e}, Michael Garcia-Ortiz, and David Filliat.
\newblock Continual state representation learning for reinforcement learning
  using generative replay.
\newblock \emph{NeurIPS}, 2018.

\bibitem[Castro et~al.(2018)Castro, Mar{\'\i}n-Jim{\'e}nez, Guil, Schmid, and
  Alahari]{Castro2018end2end}
Francisco~M Castro, Manuel~J Mar{\'\i}n-Jim{\'e}nez, Nicol{\'a}s Guil, Cordelia
  Schmid, and Karteek Alahari.
\newblock End-to-end incremental learning.
\newblock In \emph{ECCV}, 2018.

\bibitem[Chaudhry et~al.(2018)Chaudhry, Dokania, Ajanthan, and
  Torr]{chaudhry2018riemannian}
Arslan Chaudhry, Puneet~K Dokania, Thalaiyasingam Ajanthan, and Philip~HS Torr.
\newblock Riemannian walk for incremental learning: Understanding forgetting
  and intransigence.
\newblock In \emph{ECCV}, 2018.

\bibitem[Chaudhry et~al.(2019)Chaudhry, Ranzato, Rohrbach, and Elhoseiny]{AGEM}
Arslan Chaudhry, Marc’Aurelio Ranzato, Marcus Rohrbach, and Mohamed
  Elhoseiny.
\newblock Efficient lifelong learning with a-gem.
\newblock In \emph{ICLR}, 2019.

\bibitem[Dhar et~al.(2018)Dhar, Singh, Peng, Wu, and Chellappa]{dhar2018lwm}
Prithviraj Dhar, Rajat~Vikram Singh, Kuan-Chuan Peng, Ziyan Wu, and Rama
  Chellappa.
\newblock Learning without memorizing.
\newblock \emph{arXiv preprint arXiv:1811.08051}, 2018.

\bibitem[Girshick(2015)]{girshick2015fast}
Ross Girshick.
\newblock Fast r-cnn.
\newblock In \emph{ICCV}, 2015.

\bibitem[Goodfellow et~al.(2014{\natexlab{a}})Goodfellow, Pouget-Abadie, Mirza,
  Xu, Warde-Farley, Ozair, Courville, and Bengio]{goodfellow2014generative}
Ian Goodfellow, Jean Pouget-Abadie, Mehdi Mirza, Bing Xu, David Warde-Farley,
  Sherjil Ozair, Aaron Courville, and Yoshua Bengio.
\newblock Generative adversarial nets.
\newblock In \emph{NeurIPS}, 2014{\natexlab{a}}.

\bibitem[Goodfellow et~al.(2014{\natexlab{b}})Goodfellow, Mirza, Xiao,
  Courville, and Bengio]{goodfellow2013empirical}
Ian~J Goodfellow, Mehdi Mirza, Da~Xiao, Aaron Courville, and Yoshua Bengio.
\newblock An empirical investigation of catastrophic forgetting in
  gradient-based neural networks.
\newblock \emph{ICLR}, 2014{\natexlab{b}}.

\bibitem[Han et~al.(2015)Han, Pool, Tran, and Dally]{han2015weights}
Song Han, Jeff Pool, John Tran, and William Dally.
\newblock Learning both weights and connections for efficient neural network.
\newblock In \emph{NeurIPS}, 2015.

\bibitem[Han et~al.(2016)Han, Pool, Narang, Mao, Gong, Tang, Elsen, Vajda,
  Paluri, Tran, et~al.]{han2016dsd}
Song Han, Jeff Pool, Sharan Narang, Huizi Mao, Enhao Gong, Shijian Tang, Erich
  Elsen, Peter Vajda, Manohar Paluri, John Tran, et~al.
\newblock Dsd: Dense-sparse-dense training for deep neural networks.
\newblock \emph{ICLR}, 2016.

\bibitem[He et~al.(2016)He, Zhang, Ren, and Sun]{he2016deep}
Kaiming He, Xiangyu Zhang, Shaoqing Ren, and Jian Sun.
\newblock Deep residual learning for image recognition.
\newblock In \emph{CVPR}, 2016.

\bibitem[Hinton et~al.(2015)Hinton, Vinyals, and Dean]{hinton2015distilling}
Geoffrey Hinton, Oriol Vinyals, and Jeff Dean.
\newblock Distilling the knowledge in a neural network.
\newblock \emph{arXiv preprint arXiv:1503.02531}, 2015.

\bibitem[Hou et~al.(2018)Hou, Pan, Change~Loy, Wang, and
  Lin]{hou2018retrospection}
Saihui Hou, Xinyu Pan, Chen Change~Loy, Zilei Wang, and Dahua Lin.
\newblock Lifelong learning via progressive distillation and retrospection.
\newblock In \emph{ECCV}, 2018.

\bibitem[Jaderberg et~al.(2014)Jaderberg, Vedaldi, and
  Zisserman]{jaderberg2014speeding}
Max Jaderberg, Andrea Vedaldi, and Andrew Zisserman.
\newblock Speeding up convolutional neural networks with low rank expansions.
\newblock \emph{BMVC}, 2014.

\bibitem[Javed and Shafait(2018)]{javed2018revisiting}
Khurram Javed and Faisal Shafait.
\newblock Revisiting distillation and incremental classifier learning.
\newblock In \emph{ACCV}, 2018.

\bibitem[Kirkpatrick et~al.(2017)Kirkpatrick, Pascanu, Rabinowitz, Veness,
  Desjardins, Rusu, Milan, Quan, Ramalho, Grabska-Barwinska,
  et~al.]{kirkpatrick2017ewc}
James Kirkpatrick, Razvan Pascanu, Neil Rabinowitz, Joel Veness, Guillaume
  Desjardins, Andrei~A Rusu, Kieran Milan, John Quan, Tiago Ramalho, Agnieszka
  Grabska-Barwinska, et~al.
\newblock Overcoming catastrophic forgetting in neural networks.
\newblock \emph{PNAS}, 2017.

\bibitem[Krizhevsky(2009)]{krizhevsky2009cifar}
Alex Krizhevsky.
\newblock Learning multiple layers of features from tiny images.
\newblock Technical report, 2009.

\bibitem[Krizhevsky et~al.(2012)Krizhevsky, Sutskever, and
  Hinton]{krizhevsky2012imagenet}
Alex Krizhevsky, Ilya Sutskever, and Geoffrey~E Hinton.
\newblock Imagenet classification with deep convolutional neural networks.
\newblock In \emph{NeurIPS}, 2012.

\bibitem[Li et~al.(2018)Li, Li, Ding, Yang, Hu, Chen, and
  Gao]{li2018supportnet}
Yu~Li, Zhongxiao Li, Lizhong Ding, Peng Yang, Yuhui Hu, Wei Chen, and Xin Gao.
\newblock Supportnet: solving catastrophic forgetting in class incremental
  learning with support data.
\newblock \emph{arXiv preprint arXiv:1806.02942}, 2018.

\bibitem[Li and Hoiem(2018)]{li2018lwf}
Zhizhong Li and Derek Hoiem.
\newblock Learning without forgetting.
\newblock \emph{TPAMI}, 2018.

\bibitem[Liu et~al.(2017)Liu, Li, Shen, Huang, Yan, and
  Zhang]{liu2017inference}
Zhuang Liu, Jianguo Li, Zhiqiang Shen, Gao Huang, Shoumeng Yan, and Changshui
  Zhang.
\newblock Learning efficient convolutional networks through network slimming.
\newblock In \emph{ICCV}, 2017.

\bibitem[Long et~al.(2015)Long, Shelhamer, and Darrell]{long2015fully}
Jonathan Long, Evan Shelhamer, and Trevor Darrell.
\newblock Fully convolutional networks for semantic segmentation.
\newblock In \emph{CVPR}, 2015.

\bibitem[Lopez-Paz et~al.(2017)]{lopez2017gem}
David Lopez-Paz et~al.
\newblock Gradient episodic memory for continual learning.
\newblock In \emph{NeurIPS}, 2017.

\bibitem[Mallya and Lazebnik(2018)]{mallya2018packnet}
Arun Mallya and Svetlana Lazebnik.
\newblock Packnet: Adding multiple tasks to a single network by iterative
  pruning.
\newblock In \emph{CVPR}, 2018.

\bibitem[Mallya et~al.(2018)Mallya, Davis, and Lazebnik]{mallya2018piggyback}
Arun Mallya, Dillon Davis, and Svetlana Lazebnik.
\newblock Piggyback: Adapting a single network to multiple tasks by learning to
  mask weights.
\newblock In \emph{ECCV}, 2018.

\bibitem[McCloskey and Cohen(1989)]{mccloskey1989catastrophic}
Michael McCloskey and Neal~J Cohen.
\newblock Catastrophic interference in connectionist networks: The sequential
  learning problem.
\newblock In \emph{Psychology of learning and motivation}. 1989.

\bibitem[Mensink et~al.(2012)Mensink, Verbeek, Perronnin, and
  Csurka]{mensink2012ncm}
Thomas Mensink, Jakob Verbeek, Florent Perronnin, and Gabriela Csurka.
\newblock Metric learning for large scale image classification: Generalizing to
  new classes at near-zero cost.
\newblock In \emph{ECCV}. 2012.

\bibitem[Molchanov et~al.(2016)Molchanov, Tyree, Karras, Aila, and
  Kautz]{molchanov2016pruning}
Pavlo Molchanov, Stephen Tyree, Tero Karras, Timo Aila, and Jan Kautz.
\newblock Pruning convolutional neural networks for resource efficient
  inference.
\newblock \emph{ICLR}, 2016.

\bibitem[Ostapenko et~al.(2019)Ostapenko, Puscas, Klein, Jahnichen, and
  Nabi]{ostapenko2019ltr}
Oleksiy Ostapenko, Mihai Puscas, Tassilo Klein, Patrick Jahnichen, and Moin
  Nabi.
\newblock Learning to remember: A synaptic plasticity driven framework for
  continual learning.
\newblock In \emph{CVPR}, 2019.

\bibitem[Rebuffi et~al.(2017)Rebuffi, Kolesnikov, Sperl, and
  Lampert]{rebuffi2017icarl}
Sylvestre-Alvise Rebuffi, Alexander Kolesnikov, Georg Sperl, and Christoph~H
  Lampert.
\newblock icarl: Incremental classifier and representation learning.
\newblock In \emph{CVPR}, 2017.

\bibitem[Russakovsky et~al.(2015)Russakovsky, Deng, Su, Krause, Satheesh, Ma,
  Huang, Karpathy, Khosla, Bernstein, et~al.]{russakovsky2015imagenet}
Olga Russakovsky, Jia Deng, Hao Su, Jonathan Krause, Sanjeev Satheesh, Sean Ma,
  Zhiheng Huang, Andrej Karpathy, Aditya Khosla, Michael Bernstein, et~al.
\newblock Imagenet large scale visual recognition challenge.
\newblock \emph{IJCV}, 2015.

\bibitem[Selvaraju et~al.(2017)Selvaraju, Cogswell, Das, Vedantam, Parikh, and
  Batra]{selvaraju2017grad}
Ramprasaath~R Selvaraju, Michael Cogswell, Abhishek Das, Ramakrishna Vedantam,
  Devi Parikh, and Dhruv Batra.
\newblock Grad-cam: Visual explanations from deep networks via gradient-based
  localization.
\newblock In \emph{ICCV}, 2017.

\bibitem[Shmelkov et~al.(2017)Shmelkov, Schmid, and
  Alahari]{shmelkov2017object}
Konstantin Shmelkov, Cordelia Schmid, and Karteek Alahari.
\newblock Incremental learning of object detectors without catastrophic
  forgetting.
\newblock In \emph{CVPR}, 2017.

\bibitem[Wu et~al.(2019)Wu, Chen, Wang, Ye, Liu, Guo, and Fu]{wu2019BiC}
Yue Wu, Yinpeng Chen, Lijuan Wang, Yuancheng Ye, Zicheng Liu, Yandong Guo, and
  Yun Fu.
\newblock Large scale incremental learning.
\newblock In \emph{CVPR}, 2019.

\bibitem[Yu et~al.(2018)Yu, Li, Chen, Lai, Morariu, Han, Gao, Lin, and
  Davis]{yu2018nisp}
Ruichi Yu, Ang Li, Chun-Fu Chen, Jui-Hsin Lai, Vlad~I Morariu, Xintong Han,
  Mingfei Gao, Ching-Yung Lin, and Larry~S Davis.
\newblock Nisp: Pruning networks using neuron importance score propagation.
\newblock In \emph{CVPR}, 2018.

\bibitem[Zhao et~al.(2017)Zhao, Shi, Qi, Wang, and Jia]{zhao2017pyramid}
Hengshuang Zhao, Jianping Shi, Xiaojuan Qi, Xiaogang Wang, and Jiaya Jia.
\newblock Pyramid scene parsing network.
\newblock In \emph{CVPR}, 2017.

\end{thebibliography}
\end{document}